\documentclass[runningheads]{llncs}

\usepackage{eccv}

\usepackage{eccvabbrv}

\usepackage{graphicx}
\usepackage{booktabs}

\usepackage[accsupp]{axessibility}  
\usepackage{bm}
\usepackage{amssymb}
\usepackage{makecell}

\usepackage{hyperref}

\usepackage{orcidlink}

\begin{document}

\title{Towards Unsupervised Eye-Region Segmentation for Eye Tracking} 


\author{Jiangfan Deng\textsuperscript{*} \and
Zhuang Jia\textsuperscript{*} \and
Zhaoxue Wang\textsuperscript{*} \and
Xiang Long \and
Daniel K. Du}

\authorrunning{J. Deng et al.} 

\institute{ByteDance Inc.}

\maketitle

\renewcommand{\thefootnote}{}
\footnotetext{* Equal contribution}

\begin{abstract}
  Finding the eye and parsing out the parts (e.g. pupil and iris) is a key prerequisite for image-based eye tracking,
  which has become an indispensable module in today's head-mounted VR/AR devices.
  However, a typical route for training a segmenter requires tedious hand-labeling.
  In this work, we explore an unsupervised way. 
  First, we utilize priors of human eye and extract signals from the image to establish rough clues indicating the eye-region structure. 
  Upon these sparse and noisy clues, a segmentation network is trained to gradually identify the precise area for each part. 
  To achieve accurate parsing of the eye-region, 
  we first leverage the pretrained foundation model \textit{Segment Anything} (SAM) in an automatic way to refine the eye indications. 
  Then, the learning process is designed in an end-to-end manner following progressive and prior-aware principle. 
  Experiments show that our unsupervised approach can easily achieve 90\% (the pupil and iris) and 85\% (the whole eye-region) of the performances under supervised learning.
\end{abstract}

\section{Introduction}
\label{sec:intro}
\begin{figure}[h]
\begin{minipage}[]{0.33\textwidth}
  \centering
  \includegraphics[width=1.\textwidth]{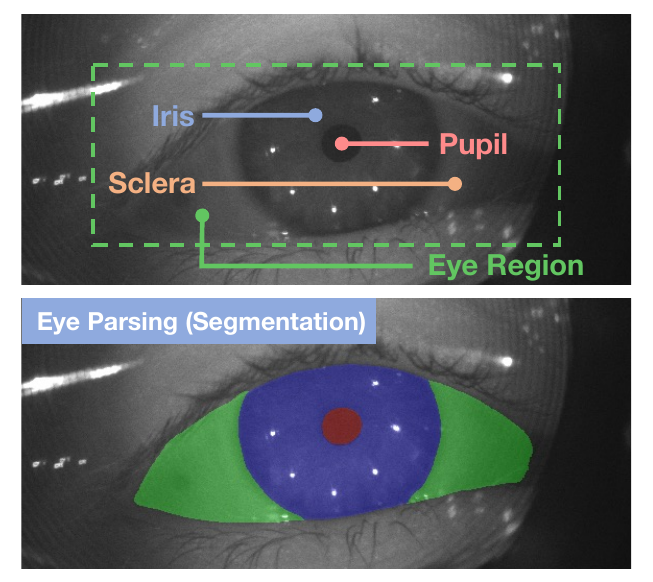}
\end{minipage}
\hskip 4pt plus 0fil
\begin{minipage}[]{0.67\textwidth}
  \centering
  \caption{{\bf{Demonstration of our task.}}
    The upper image is an example of the photo captured by the near-infrared camera pointing to the eye. 
    Each part of the eye (the pupil, iris and the sclera) is marked using an individual color. 
    Our goal is paring the eye-region: finding out the area of pupil, iris and the whole region of eye from the image, 
    or more specifically, doing a semantic segmentation of the eye and its parts (as shown in the lower image). 
    We seek to accomplish this task through an unsupervised way. 
  }
  \label{fig:task}
\end{minipage}
\vspace{-1\baselineskip}
\end{figure}

Eye tracking technology is becoming increasingly important in recent years. 
When integrated into a VR/AR device, it can yield rich insights into the visual processes of human being~\cite{fuhl2018automatic,geisler2017saliency}, shedding light on the user's intentions and actions.  
These valuable information can be widely adopted in many fields such as gaze-dependent rendering~\cite{patney2016towards,cholewiak2017chromablur,levoy1990gaze,mantiuk2013gaze}, medical diagnosis~\cite{wang2022follow,saab2021observational,waisberg2024apple}, remote support~\cite{gupta2016you,yao2018visualizing} and so on. 
It also holds the potential to revolutionize the way of human-machine interaction~\cite{hutchinson1989human,fuhl2021perception,drewes2010eye,jungwirth2018eyes}. 
In head-mounted devices, eye tracking often relies on images captured by a near-infrared camera pointing to the eye.
Therefore, an important preliminary step is to identify the eye and its parts like the pupil and iris (Fig.~\ref{fig:task}). 
Generally, a typical way to accomplish this task is to collect and annotate a large amount of images and then train a segmentation model.
However, manually labeling the pixel-wise mask is a labor-consuming work. 
When facing rapid updates of the hardware prototype, this approach proves to be highly inefficient.
%
%

In this work, we explore a pathway to do the eye-region segmentation in an unsupervised manner.
The possibility stems from two facts:
1) priors of human eye are strong and stable and 
2) a segmentation network with dense prediction can be trained by sparse annotations.
Following these principles, we design our method concisely. 
First, there is an observation that the brightness on the image increases step by step from the pupil to the iris and then to the sclera.
So after locating a point within the pupil region,
gradient on each pixel is checked along a radial direction to probe the potential boundary for pupil and iris,
upon which we can roughly ``indicate'' the area of these two parts. 
Then we employ a segmentation network~\cite{guo2018review} trying to learn their dense masks through the sparse indications above. 
After being able to identify boundaries of pupil and iris, the segmenter is further guided to explore the whole eye-region. 
First, we use the iris prediction and the local standard deviation of gradients on each image to produce initial indications for the eye.  
Then, the pretrained segmentation model \textit{Segment Anything} (SAM)~\cite{kirillov2023segment} is leveraged automatically, 
helping the eye indicator restraining errors and mining more positive samples. 
Since the auto-generated indications contain considerable errors, 
we design the learning process based on a progressive and prior-aware principle, 
which effectively resist noise in the training signal. 
The entire pipeline runs in an end-to-end scheme and ends up with a model that simultaneously segments the eye, iris, and pupil, 
with the acquisition process being totally unsupervised. 

We conduct comprehensive and rigorous evaluation of our method across multiple datasets. 
Thanks to the insensitive hyper-parameters and general design,
the algorithm can easily adapt to different domains (such as images captured by various head-mounted devices, or from diverse angles).
Experimental results indicate that our method can achieve 90\% of the performance under fully-supervised learning for pupil and iris, and 85\% for the more challenging eye-region.

\section{Related Work}
\label{sec:related}

\noindent\textbf{Eye-Region Parsing.}
Eye-region parsing is an essential step for eye tracking. 
Recently, there have been numerous efforts dedicated to this task, exploring improvements in both accuracy and efficiency. 
Early works usually extract low-level features from the image and apply semantic priors of the eye-region. 
Among them, many approaches are solely focused on finding pupil center~\cite{fuhl2015excuse,li2005starburst,fuhl2017fast,santini2018pure,poulopoulos2021pupiltan,fuhl2018bore} or pupil area~\cite{fuhl2016else,fuhl20211000}.
For example, ExCuSe~\cite{fuhl2015excuse} computes Canny edges~\cite{canny1986computational} on the image and locate the pupil by selecting the curves progressively. 
Starburst~\cite{li2005starburst} uses derivative of pixels along radial directions to detect the boundary of pupil, searching the pupil center via several  interations. 
ElSe~\cite{fuhl2016else} starts by extracting and filtering edges on the image and then finds out pupil by fitting and selecting ellipse. 
In~\cite{fuhl2017fast}, the authors try to detect the more challenging eyelids through approximation with polynomials.
The main drawback of these methods lies in their heavy reliance on rules and thresholds. 
With the development of convolutional neural networks~\cite{lecun1998gradient}, many model-based methods have emerged. 
In the context of deep learning, parsing the eye-region is essentially a task of semantic segmentation. 
In~\cite{kim2019eye}, the authors propose a muti-class segmentation network with encode-decoder architecture. 
Then a heuristic filtering module is adopted to refine the outputs. 
In RITnet~\cite{chaudhary2019ritnet}, domain-specific augmentations and boundary aware loss functions are used to optimize the training, 
making the lightweight model achieving precise eye segmentation. 
EllSeg~\cite{kothari2021ellseg} utilizes the geometry assumption of pupil and iris, guiding the network to segment the full but not entirely visible ellipses. 
In our work, we embrace the modeling paradigm of semantic segmentation while steer clear of manual labeling. 
We seek to leverage low-level features and prior knowledge to inspire the segmentation network, 
while avoiding intricate rules and excessive hyper-parameters at the same time. 

\noindent\textbf{Unsupervised/Semi-supervised Segmentation.}
The task of visual segmentation needs pixel-wise annotations, which is labor-intensive. 
Recently, segmentation based on unlabeled data (unsupervised learning) or only a few part of labeled data (semi-supervised learning) is attracting attentions. 
Under the ``unsupervised'' setting, 
IIC~\cite{ji2019invariant} maximize mutual information of patch-level cluster assignments between an image and its augmentations to acquire pixel-level clustering. 
PiCIE~\cite{cho2021picie} uses invariance to photometric effects and equivariance to geometric transformations in optimizing the segmentation. 
In~\cite{hamilton2022unsupervised}, existing pretrained self-supervised features are refined to enhance the performance. 
In our task, since the near-infrared eye image is particularly specific, 
the existing unsupervised approaches cannot perform optimally due to their inability to utilize prior knowledges.  
Under the ``semi-supervised'' setting, many works apply the idea of consistency regularization to this field.  
In~\cite{ouali2020semi}, a cross-consistency training method is proposed, 
in which the perbutations of the encoder outputs are used to enforce the invariance of prediction. 
\cite{lai2021semi} maintains the context-aware consistency under different environments via diverse cropping and a directional contrastive loss function. 
In~\cite{cheng2022pointly}, it is experimentally proved that the dense mask of an object can be trained by sparse points, 
which is an important conclusion we make use of in this paper. 
There are also attempts to mitigate the annotation reliance in the task of eye segmentation.  
For example, \cite{chaudhary2021semi} makes a semi-supervised learning framework for eye segmentation and uses spatially varying augmentations to optimize the results. 
In~\cite{cai2021landmark}, an ingenious landmark-aware method of self-supervised eye segmentation is proposed. 
However, since it depends on the keypoint pictorial representations from labeled data, the setting is not exactly unsupervised. 
In our work, by leveraging low-level signals from the image and the semantic ability of segmentation models, 
we achieve acceptable segmentation results under the strict ``unsupervised'' setting.

\section{Methodology}
\label{sec:method}
\begin{figure}[t]
  \centering
  \includegraphics[width=1.\linewidth]{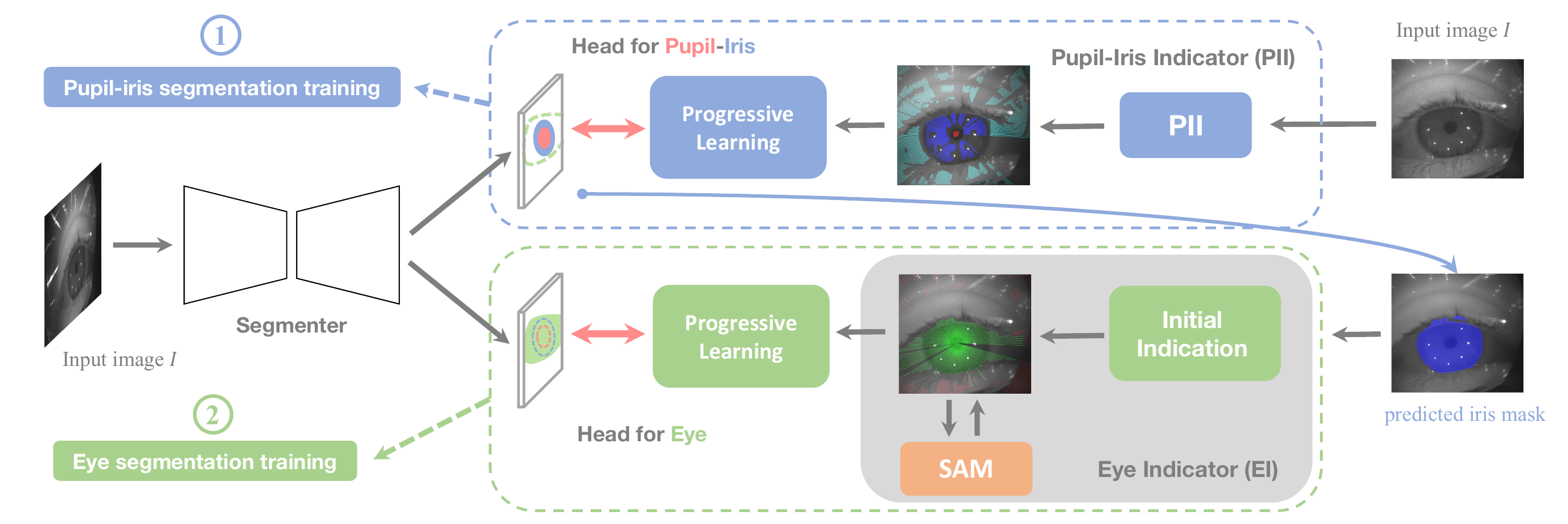}
  \vspace{-1.5\baselineskip}
  \caption{{\bf{Overview of our method.}}
           The segmentation network has two heads: one for pupil/iris and the other for eye-region.
           Training of pupil/iris segmentation is first activated using indications generated by PII.
           Then the eye segmentation head is unlocked to update, using indications from EI, which utilizes the pupil/iris prediction.
           After that, both tasks are running simultaneously. 
           During the training process, a progressive learning module is used for each task, resisting the noise and refining the outputs.
          }
\label{fig:overview}
\end{figure}

The overall idea is first seeking for clues about objects and then use them to supervise the segmentation training. 
Fig.~\ref{fig:overview} summarizes the entire process. 
First, an image is sent into a module named Pupil-Iris Indicator (PII), generating sparse indications of the two parts.
During this time, a segmentation network is activated, training upon these coarse information.
Since the dense predictions can be inspired by sparse annotations, 
the network will gradually learn to identify the boundaries of pupil and iris, 
after which we further explore the whole area of eye.
Continue in Fig.~\ref{fig:overview}, the image with its  iris prediction are sent into another module named Eye Indicator (EI).
Following similar principles as PII, indications of the eye-region are generated 
and the segmentation model opens another prediction head, learning to segment the eye.
After the activation of both heads, the training tasks of pupil/iris segmentation and eye segmentation work simultaneously. 
%
%

The core signal we exploit in both indicators is the \textit{gradient} on the image, 
which provides rich information so that the visual structure of eye-region can largely be revealed. 
In PII and EI, we encapsulate a series of gradient-aware operations, making them unified and general. 
To handle the more challenging task of distinguishing eyelids, 
we unleash the potential of the \textit{Segment Anything} (SAM)~\cite{kirillov2023segment} and use it in an automatic way. 
Moreover, the training process is launched obeying a progressive and prior-aware principle, 
which effectively resists noise in the rough training signals. 
In the rest of this section, we go into details of the pupil-iris indicator (PII), the eye indicator (EI) and the learning process.

\subsection{Pupil-Iris Indicator}
\label{sec:pii}
%
%
\begin{figure}[t]
  \centering
  \includegraphics[width=1.\linewidth]{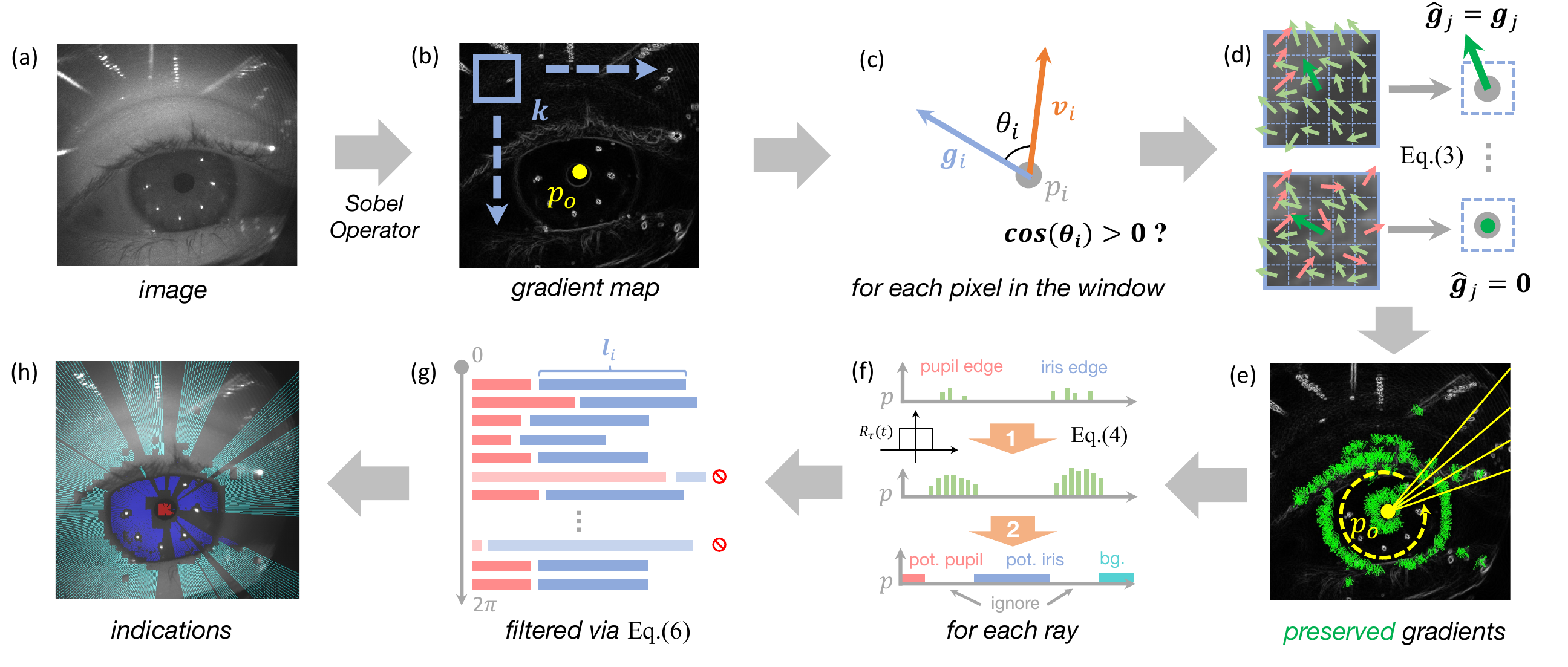}
  \vspace{-1.5\baselineskip}
  \caption{{\bf{Pupil-Iris Indicator.}}
            First, gradients on the input image are computed using Sobel operator (a,b).
            Then, Eq.(\ref{eq:cos}) and Eq.(\ref{eq:gdf}) are adopted to filter gradients within each window (b,c,d).
            After that, an indicating process based on a set of rays originating from $p_o$ is a applied (e,f,g), 
            generating sparse indications for pupil and iris (h).
          }
\label{fig:pii}
\end{figure}

In a near-infrared image of human eye, there is a stable observation that the brightness increases step by step from the pupil to the iris and then to the sclera. 
This phenomenon can be well described in the gradient space: 
gradients on the boundary of pupil and iris are likely to exhibit a radial orientation from inside the pupil/iris to the outside. 
In PII, we first roughly locate one point inside the pupil 
and then filter gradient on each pixel along the rays starting from this point, probing the pupil and iris boundary. 

\noindent\textbf{Find a Point in Pupil.}
A simple method is used to find an arbitrary point within the pupil: 
First, the image is lightened adaptively to avoid the influence of mid- and light-areas. 
Then we use Haar-like filter with different radii~\cite{swirski2012robust} to extract the candidate pupil locations by finding top-$k$ largest responses.
After simply filtering the extreme area and shape of their connected components, 
the darkest location left is considered as the expected point. 

\noindent\textbf{Radial Indication.}
Fig.~\ref{fig:pii} demonstrates the process. 
Formally, Let $I\in\mathbb{R}^{w\times{h}}$ be the input image (with single channel) and $p_o\in{I}$ be the point within pupil.
We first compute gradients $\bm{G}\in\mathbb{R}^{w\times{h}\times{2}}$ of the image using Sobel operator~\cite{sobel2022sobel}:
\begin{equation}
  \bm{G} = \mathbf{Sobel}(I),
\label{eq:grad}
\end{equation}
where each position on $\bm{G}$ stores a gradient vector $\bm{g}_i\in\mathbb{R}^2$. 
For every pixel $p_i\in{I}$, another vector $\bm{v}_i$ from $p_o$ to $p_i$ can also be acquired.
Therefore, the observation mentioned above has a more clear description: 
if $p_i$ lies on the boundary of pupil or iris, the condition bellow would largely be satisfied:
\begin{equation}
  \cos\theta_i=\frac{\bm{g}_i\cdot{\bm{v}_i}}{\|\bm{g}_i\|\|\bm{v}_i\|}\textgreater{0}, 
\label{eq:cos}
\end{equation}
where $\theta_i$ is the angle spanned by $\bm{g}_i$ and $\bm{v}_i$.
Now we apply this rule throughout the image. 
As shown in Fig.~\ref{fig:pii}-(b), a sliding window $k$ scans the image plane.
For each pixel $p_j$, its gradient $\bm{g}_j$ will be preserved if more than $r_{th}=80\%$ of gradients in the window $k_j$ centered at $p_j$ satisfying Eq.(\ref{eq:cos}), otherwise, it will be set to $\bm{0}$.
This process (Fig.~\ref{fig:pii}-(c) and (d)) can be written as:
\begin{equation}
  \hat{\bm{g}}_j={\bm{g}_j}\cdot{\mathbf{1}_{\mathbb{R_+}}((\frac{1}{|k_j|}\sum_{p_i\in{k_j}}\mathbf{1}_{\mathbb{R_+}}(\cos\theta_i))-r_{th})},
\label{eq:gdf}
\end{equation}
where $\mathbf{1}_{\mathbb{R}_+}(\cdot)$ is the indicator function that outputs 0 or 1.
As Fig.~\ref{eq:gdf}-(e) shows, the preserved gradients can roughly outline the pupil and iris, although with some noise.

Next we start to generate indications based on the filtered gradients above.  
At the beginning, a set of rays originating from $p_o$ are posed, covering the entire angular range of $[0,2\pi)$ (Fig.~\ref{fig:pii}-(e)).
Basically, each ray will first pass through the boundary of pupil and then the boundary of iris.
We leverage a single-dimensional convolution along each ray to apply this assumption: 
for the $i$-th ray $\bm{r}_i$, a rectangular pulse function is employed to convolve with gradient magnitudes on it (the first step in Fig.~\ref{fig:pii}-(f)): 
\begin{equation}
r^{\ast}_i(t)=\sum_{m=-\infty}^{\infty}r_i(t-m)\cdot{R_{\tau}(m)}, \
t\in\{1,...,T\},
\label{eq:conv}
\end{equation}
where $T$ is the length of the ray and $R_{\tau}(\cdot)$ refers to the rectangular pulse function with the width $\tau$: 
\begin{equation}
R_{\tau}(t)=\left\{\begin{array}{lrl} 1, & & \mbox{if} -\frac{\tau}{2}<t<\frac{\tau}{2} \\
              0, & & \mbox{elsewise}
              \end{array}\right.
\label{eq:pulse}
\end{equation}
Naturally, we assume the first line segment with non-zero values is the pupil boundary and the second one is the iris boundary.
Therefore, we can sequentially mark the {\color{red}potential pupil} (line segment before the pupil boundary) and {\color{blue}potential iris} (line segment between the pupil boundary and the iris boundary) along the ray $\bm{r}_i$: 
as show in the second step of Fig.~\ref{fig:pii}-(f), 
pixels on each line segment are indicated as either pupil or iris and those beyond the iris boundary are indicated as background, 
while the rest pixels on the ray will be ignored.  
As previously stated, the labeling process is inevitably flawed. 
We mitigate the noise by using statistics from all rays in $[0,2\pi)$.  
Fig.~\ref{fig:pii}-(g) exhibits the process. 
First, computing average values $l_{ave}$ and standard deviations $l_{std}$ for the length of all line segments for potential iris:
\begin{equation}
l_{ave}=\frac{1}{N}\sum_{i=1}^{N}l_i,\quad l_{std}=\sqrt{\frac{1}{N}\sum_{i=1}^{N}(l_i - l_{ave})^2}. 
\label{eq:mstd}
\end{equation}
Then, for each ray, labels on it will be ignored if the length $l_i$ of iris line segment on the ray does not satisify $|l_i-l_{ave}|\leqslant{l_{std}}$. 
The map of pupil-iris indications is finally obtained as shown in Fig.~\ref{fig:pii}-(h). 

\subsection{Eye Indicator}
\label{sec:ei}
The segmenter will first be trained on pupil and iris segmentation task using indications generated from PII (training process will be described later). 
Then, the Eye Indicator (EI) proceeds to explore the entire eye-region based on relatively high-quality iris prediction, 
enabling the eye segmentation training.

\begin{figure}[t]
  \centering
  \includegraphics[width=1.\linewidth]{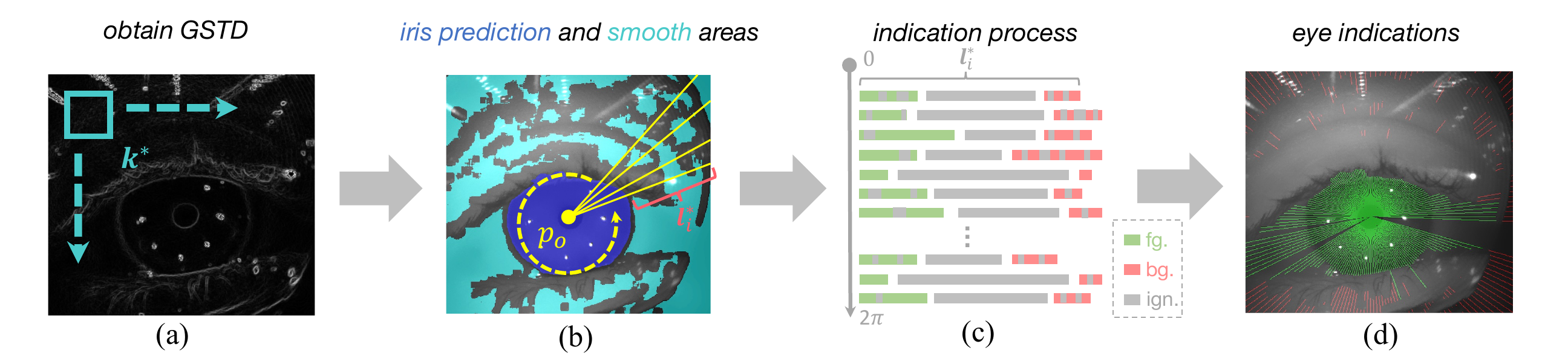}
  \vspace{-1.5\baselineskip}
  \caption{{\bf{Initial Indication of eye.}}
           (a) Compute GSTD on the gradient map.
           (b) Smooth areas on the image (marked in light blue).
           (c) Indicate the eye.
           (d) Output indications: \textit{green}: eye (foreground), \textit{red}: background, \textit{others}: ignored.
          }
\label{fig:eri}
\end{figure}

\vspace{1em}
\noindent\textbf{Initial Indication.}
Empirically, the eye's sclera is likely to be smooth. 
We estimate the smoothness using the standard deviation of gradient magnitudes (denoted as GSTD). 
Like the operation in PII, a sliding window $k^{*}$ is used to scan the gradient map (Fig.~\ref{fig:eri}-(a)). 
For each pixel $p_i$, we compute the GSTD in the window centered at $p_i$.  
The pixel will be marked as \textit{smooth} if the GSTD is less than a threshold $th_{std}$, 
otherwise it will be marked as \textit{rough} (Fig.~\ref{fig:eri}-(b)).
Then we reuse the set of rays in Sec.~\ref{sec:pii}.  
In all likelihood, each ray will first pass through the pupil and iris and then the sclera. 
Since the iris prediction is available (dark blue area in Fig.~\ref{fig:eri}-(b)), 
we indicate the sclera region along the ray following a simple rule (Fig.~\ref{fig:eri}-(c)): 
First, the area of (predicted) pupil and iris is naturally labeled as foreground.
Let $l^*_i$ be the part of the $i$-th ray after removing the pupil and iris region.
Starting from the endpoint of iris, 
the first 30\% of \textit{smooth} line segments on $l^*_i$ are indicated as foreground and 
the last 30\% \textit{smooth} line segments are indicated as background, while the rest are ignored. 
Therefore, we get the initial eye indications (Fig.~\ref{fig:eri}-(d)).
%
%

\begin{figure}[t]
  \centering
  \includegraphics[width=1.\linewidth]{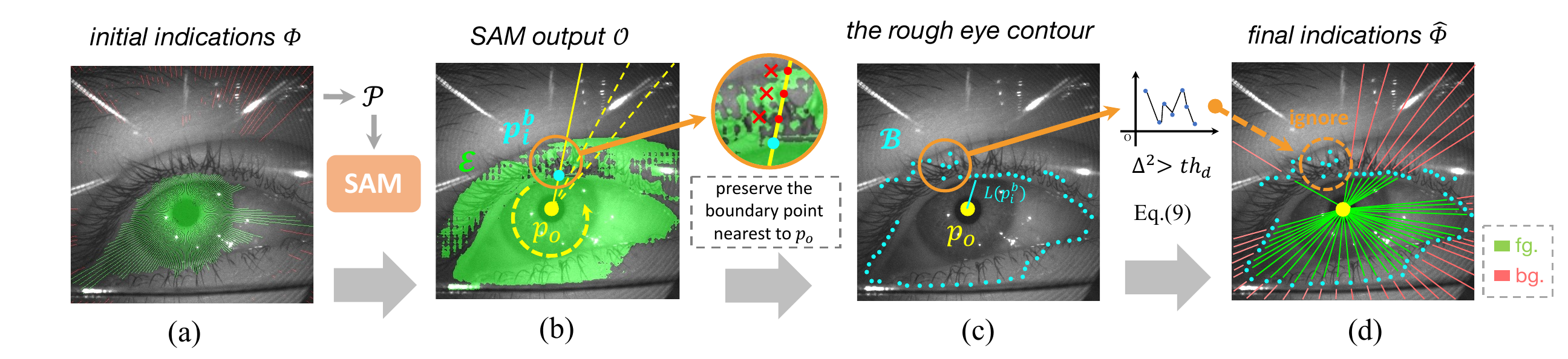}
  \vspace{-1.5\baselineskip}
  \caption{{\bf{Indication Refinement with SAM.}}
           The initial indications $\Phi$ are used to generate point-prompts $\mathcal{P}$ to interact with SAM (a). 
           Based on the SAM output $\mathcal{O}$, the set $\mathcal{B}$ of points on the rough contour of eye is acquired (b,c). 
           Then, contour sections which are not smooth will be filtered by computing second-order derivatives (c,d).
           The final ``reliable but incomplete'' eye indications $\hat{\Phi}$ is obtained (d). 
          }
\label{fig:sampm}
\end{figure}

\vspace{1em}
\noindent\textbf{Indication Refinement with SAM.}
Comparing to pupil/iris, variations of eye contours are diverse and intricate, 
leading to incorrect labeling and missing of critical areas while indicating the eye. 
To remedy this issue, we import the pretrained segmentation model \textit{Segment Anything} (SAM)~\cite{kirillov2023segment}, 
leveraging its semantic abilities to restrain errors and find more positive samples. 
Fig.~\ref{fig:sampm} shows the process. 
Naturally, the initial eye indications (denoted as $\Phi$) are used for prompt generation. 
We first divide the image plane into $n\times{n}$ grids (we fix $n$ to 10 in our method). 
In each grid, one of the dominated (positive or negative) samples from the corresponding grid in $\Phi$ is randomly selected as a (positive or negative) prompt. 
The set of all prompts on the image can be written as:
\begin{equation}
\mathcal{P}=\{p_1^-, \varnothing, p_3^+, p_4^-, ..., p_{n\times{n}}^+\}, 
\label{eq:prompts}
\end{equation}
where $\varnothing$ denotes the possible prompt-absence of the grid because all pixels within it are ignored in $\Phi$. 
Then we use $\mathcal{P}$ to interact with SAM and obtain the segmentation outputs $\mathcal{O}\in\{0,1\}^{w\times{h}}$ (Fig~\ref{fig:sampm}-(b)). 
Obviously, $\mathcal{O}$ cannot be directly adopted as training targets since the responses from SAM are still noisy (\cite{maquiling2023zero} provides typical cases). 
We invent a method to preserve reliable samples while disregarding the possible errors. 
%
%

As shown in Fig~\ref{fig:sampm}-(b). First, the set of rays in Sec.~\ref{sec:pii} is applied again on $\mathcal{O}$. 
Let $\mathcal{E}\subset{\mathcal{O}}$ be the positive samples (the eye-region predicted by SAM). 
Along the $i$-th ray, a point $p_i^b$ on the boundary of $\mathcal{E}$ will be acquired 
(we only preserve the point nearest to the origin $p_o$ if there is more than one boundary points). 
Therefore, all boundary points $\mathcal{B}=\{p_1^b, ..., p_N^b\}$ on the image ($N$ is the number of rays) can be roughly seemed as the eye-region contour with inevitable errors (Fig.~\ref{fig:sampm}-(c)). 
Empirically, a section of contour should be smooth if being well aligned to the eyelids, 
otherwise, it will exhibit large local variation with sudden changes. 
We use the second-order derivative to characterize this property and try to filter out the incorrect contour points (Fig.~\ref{fig:sampm}-(c),(d)). 
Assuming there is a function $r(\cdot)$ describing the eye contour on the polar coordinates with the origin located at $p_o$. 
The second-order derivative of $r(\cdot)$ at $\theta$ can be defined as: 
\begin{equation}
\frac{dr^2(\theta)}{d\theta^2}=\lim_{\delta\rightarrow0}\frac{r(\theta+\delta)-2r(\theta)+r(\theta-\delta)}{\delta^2}.
\label{eq:sodc}
\end{equation}
Since the $\theta$ is a set of discrete points $p_i^b\in\mathcal{B}$ in our task, 
we approximate Eq.(\ref{eq:sodc}) using finite difference:
\begin{equation}
\Delta^2_i = L(p_{i+1}^b)-2L(p_i^b)+L(p_{i-1}^b),\quad i=1,..., N,
\label{eq:sodd}
\end{equation}
where $L(p)$ denotes the distance of pixel $p$ from the origin $p_o$. 
Then we inspect all boundary pixels in $\mathcal{B}$ to determine whether they should be preserved. 
Specifically, for each point $p_i^b\in\mathcal{B}$, we define a neighborhood $\mathcal{N}_{\epsilon}(p_i^b)\subset{\mathcal{B}}$ along the boundary with the size $\epsilon$. 
If there exists a point $p_j^b\in\mathcal{N}_{\epsilon}(p_i^b)$ with \textit{large derivative} ($\Delta^2_j$ is greater than a threshold $th_{d}$), we delete $p_i^b$ from $\mathcal{B}$, otherwise, it will be retained. 

After the filtering process, we get a new set of points $\hat{\mathcal{B}}\subset{\mathcal{B}}$ which are regarded as the ``reliable but incomplete'' eye contours on the image. 
$\hat{\mathcal{B}}$ is then leveraged to update the initial indications $\Phi$: 
for each ray $\bm{r}_i$, if the corresponding $p_i^b$ is retained ($p_i^b\in\hat{\mathcal{B}}$), 
we mark all pixels between $p_o$ and $p_i^b$ on $\bm{r}_i$ as positive and those beyond $p_i^b$ as negative. 
The updated eye indications $\hat{\Phi}$ will be used for training (Fig.~\ref{fig:sampm}-(d)). 

\subsection{Learning Process}
\label{sec:pl}
Both the pupil-iris indicator (PII) and the eye-indicator (EI) will be intergated into a unified learning process as signal providers. 
We arrange the training scheme following a progressive rule, enabling the segmenter evolving from coarse to fine. 
First, the PII module depends on the initial point $p_o$. 
So after the training of the pupil-iris head having converged to some extent, 
we use the weighted center of the predicted pupil mask ($M_p$) as the initial point instead of the searching process in Sec.~\ref{sec:pii}:
\begin{equation}
p_o=\frac{1}{|M_p|}\sum_{p_i\in{M_p}}p_i. 
\label{eq:sodd}
\end{equation}
Second, the EI module depends on the pupil-iris predictions. 
Therefore, at the same time of updating $p_o$, we can obtain these predictions to actuate EI. 
The whole process is designed in a coordinated manner: 
starting from a pre-defined training epoch $E_{start}$, the model will be managed to inference over the entire train set. 
Following that, we open the PII and EI in parallel to update the indcations of pupil/iris and eye. 
This process will be repeated every $E_{step}$ epochs. 

Furthermore, we leverage a simple prior to resist the indication noise during training. 
As mentioned before, the luminance of pupil region is always lower than that of iris. 
Since there are labeling errors in the output of PII 
(e.g. the line segment labeled as pupil on a ray might surpass the real pupil edge),  
we design a ignoring strategy to alleviate their negative effects. 
Specifically, at each iteration, let $M_p$ and $M_{ir}$ denote the predictions of pupil and iris. 
We first compute the average luminance $C$ within the two regions respectively: 
\begin{equation}
C_p=\frac{1}{|M_p|}\sum_{i=1}^{|M_p|}c_i, \quad C_{ir}=\frac{1}{|M_{ir}|}\sum_{j=1}^{|M_{ir}|}c_j. 
\label{eq:ave_lumi}
\end{equation}
Then, a luminance threshold $th_{c}=(C_p+C_{ir})/2$ is defined. 
Since the brightness of pupil and iris pose a clear clustering effect and the gap between them is relatively stable, 
this threshold can effectively inspect possible errors in $M_p$ and $M_{ir}$. 
Therefore, in the correponding training step, we ignore pixels in $M_p$ with luminance greater than $th_{c}$ and pixels in $M_{ir}$ with luminance less than $th_{c}$.

\section{Experiment}
\label{sec:experiment}{}
To validate the feasibility and generalization of the proposed method, 
we conduct thorough experiments across multiple datasets. 
The results suggest that our approach is capable of yielding acceptable results in unsupervised scenarios without requiring meticulous tuning of hyper-parameters. 

\noindent\textbf{Datasets.}
In our experiments, we mainly use four datasets: OpenEDS-19~\cite{garbin2019openeds}, OpenEDS-20~\cite{palmero2020openeds2020}, NVGaze-AR~\cite{kim2019nvgaze} and Gaze-in-wild (GIW)~\cite{kothari2020gaze}. 
The OpenEDS-19 and OpenEDS-20 collect eye images from head-mounted VR devices and provide annotations for eye-region segmentation. 
In OpenEDS-19, there are 152 participants and 12759 labeled images in total, 
among which the train, validation and test datasets contain 8916, 2403 and 1440 images respectively. 
The OpenEDS-20 dataset includes 200 frame sequences from 74 participants. 
Then, nearly 5\% of the frames are hand-labeled with segmentation mask, 
in which 5 frames in each sequence are randomly selected for testing while the rest labeled frames are used for training. 
The NVGaze-AR and GIW do not originally provide mask labels. 
Thanks to the contribution of TEyeD~\cite{fuhl2021teyed}, we can access annotations of segmentation, landmarks and gaze information of them. 
In order to properly use these data in our task and make the experiments reasonable, 
we re-arrange the two datasets following four principles: 
(1) Sample images from the original video clips, reducing the duplicated frames by examining similar gaze directions. 
(2) Make sure the train and test set do not contain images from the same subjects (participants). 
(3) Keep only the visible part of pupil and iris (masks provided by TEyeD are full ellipse of pupil/iris). 
(4) Filter out images that do not contain visible pupil. 
Table.~\ref{tab:datasets} exhibits the information of all datasets we used.  
It should be noted that we cancel the category of \textit{sclera} and use the concept of \textit{eye} (combination of the pupil, iris and sclera) instead. 
We think this definition is more practical for the task of eye tracking. 
%
%
\begin{table}[]
\small
\centering
\begin{tabular}{l|c|c|c|c}
\toprule
                                             &  image size ($w\times{h}$)          & train       & test & annotation source \\ \hline
OpenEDS-19~\cite{garbin2019openeds}          &  $400\times{640}$    &  8916        &     1440    &   original \\
OpenEDS-20~\cite{palmero2020openeds2020}     &  $640\times{400}$    &  1605        &     1000    &  original \\
NVGaze-AR~\cite{kim2019nvgaze}               &  $640\times{480}$    &  29114       &     8173    & TEyeD~\cite{fuhl2021teyed}  \\
Gaze-in-wild~\cite{kothari2020gaze}          &  $640\times{480}$    &  77817       &     20816   & TEyeD~\cite{fuhl2021teyed} \\
\bottomrule
\end{tabular}
\vskip 5pt plus 1fil
  \caption{\small{
    \textbf{Datasets Information.}
    We use four datasets in our experiments: OpenEDS-19~\cite{garbin2019openeds}, OpenEDS-20~\cite{palmero2020openeds2020}, 
    NVGaze-AR~\cite{kim2019nvgaze} and Gaze-in-wild~\cite{kothari2020gaze}, 
    Among which the first two datasets originally provides segmentation labels 
    and annotations of the last two datasets are obtained from TEyeD~\cite{fuhl2021teyed}. 
  }}
\label{tab:datasets}
\end{table}

\vspace{-0.5cm}
\noindent\textbf{Supervised Benchmarks.}
We adopt RITnet~\cite{chaudhary2019ritnet} as the network structure in our experiments, 
which is currently the state-of-the-art in the field of eye-region segmentation. 
To adapt our task and highlight the key factors, we make simplifications on its original version to build our supervised benchmarks. 
First, the single segmentation head of three categories is reformed to double heads for pupil/iris and eye respectively as demonstrated in Fig.~\ref{fig:overview}. 
Then, we give up the gamma correction and CLAHE in data pre-processing and cancel all the training augmentations except the horizontal flipping. 
During training, only the standard cross-entropy loss (CEL) is adopted. 
Each training process is conducted on 8 GPUs and we use a total batch-size of 64 with 8 images for each GPU. 
The original image sizes are preserved in both train and test processes and each dataset is trained for 300 (for OpenEDS-20) or 100 (for OpenEDS-19, NVGaze-AR and GIW) epochs. 
We adopt the AdamW~\cite{loshchilov2017decoupled} optimizer with a initial learning rate of 0.0025. 
Then the learning rate will be reduced by 0.1 at 1/2 and 2/3 of the whole training epochs. 

\noindent\textbf{Implementation Details.}
A unified set of hyper-parameters are used across the four datasets. 
In PII (Sec.~\ref{sec:pii}), we set the size of the sliding window $k$ to $10\times{10}$ and the width $\tau$ of the rectangular pulse function $R_{\tau}(\cdot)$ to 3. 
In EI (Sec.~\ref{sec:ei}), the size of $k^{*}$ is set to $5\times{5}$, the GSTD threshold $th_{std}$ is set to 5, the difference threshold $th_d$ is set to 20 and the size $\epsilon$ of boundary neighborhood is set to 30. 
In the indication-refinement process, we adopt SAM with the heavy image encoder ViT-H~\cite{dosovitskiy2020image}. 
During training, the prediction head of eye is activated at 1/4 of the whole training scheduler ($E_{start}=0.25\times{epochs}$) and the PII and EI module are driven to update the indications for every 30 epoches ($E_{step}=30$). 
The prior-aware learning strategy also starts working at $E_{start}$. 
Other training settings are kept the same with those in the supervised benchmarks. 

\subsection{Main Results}
We make comparisions of our method with three counterparts: 
the supervised benchmark, the semi-supervised learning (SSL) approach proposed in~\cite{chaudhary2021semi} and the latest unsupervised segmentation method STEGO~\cite{hamilton2022unsupervised}. 
In SSL, we randomly select 1\% images from the train sets as labeled data and preserve all the augmentation processes in~\cite{chaudhary2021semi} during training. 
In STEGO, since all categories of our task (pupil, iris and eye) consistently appear in a single image, which constrains the potential of contrastive learning, 
we maintain the pretrained backbone frozen and only update the segmentation head. 
After the final clustering is finished, the Hungarian matching~\cite{kuhn1955hungarian} algorithm is used to assign the clusters with ground truth for evaluation. 
Table.~\ref{tab:main} demonstrate the main results. 
Comparing to the supervised benchmark, 
the proposed method can achieve 90\% of the performance for pupil/iris and 85\% of the performance for eye. 
This percentage number can even surpass 94\% for the category of pupil (in OpenEDS-19, OpenEDS-20 and NVGaze-AR), which is the most crucial part in the eye-tracking task. 
As exhibited in the second line for each dataset in Table.~\ref{tab:main}, 
the semi-supervised learning (SSL) approach shows promising performance and some results of it are approaching the supervised benchmarks. 
It is important to note that due to the constrained nature of near-infrared eye image scenarios, a small amount of labeled data (or even representations substracted from the labeled data~\cite{cai2021landmark}) can yield a significant amount of information, whereas our unsupervised setting poses a much greater challenge. 
For the unsupervised segmentation algorithm STEGO, although a much heavier transformer-based model~\cite{caron2021emerging} is adopted, 
performances decrease drastically on all categories across the four datasets, especially for the pupil (completely fail). 
In our opinion, the reason lies in the fact that images of this task are in a very specific domain 
and the concept of eye structure is too abstract without any prior knowledge. 
%
%
\begin{table}[]
\scriptsize
\centering
\begin{tabular}{l|ccc|ccc}
\toprule
                                        & Pupil   & Iris   & Eye   & Pupil  & Iris & Eye \\ \hline 
\makecell{\textbf{Method}}              & \multicolumn{3}{c|}{\textbf{OpenEDS-19}} & \multicolumn{3}{c}{\textbf{OpenEDS-20}} \\ \hline
Supervised-benchmark                    &  88.33        &  93.22        &  95.22        &  94.22        &  95.14        &  95.76           \\
SSL~\cite{chaudhary2021semi}            &  87.01        &  91.05        &  91.88        &  91.06        &  90.22        &  90.63           \\
STEGO~\cite{hamilton2022unsupervised}   &  0.18         &  44.32        &  48.40        &  0.13         &  41.25        &  29.36           \\
\textbf{Ours}                           &  83.05 (94\%) &  82.62 (89\%) &  80.48 (85\%) &  88.32 (94\%) &  87.06 (92\%) &  81.43 (85\%)    \\ \hline
                                        & \multicolumn{3}{c|}{\textbf{NVGaze-AR}}             & \multicolumn{3}{c}{\textbf{GIW}}           \\ \hline
Supervised-benchmark                    &  94.37        &  94.08        &  92.91        &  93.83        &  90.95        &  91.96           \\
SSL~\cite{chaudhary2021semi}            &  89.55        &  90.01        &  88.75        &  86.52        &  87.03        &  88.02           \\
STEGO~\cite{hamilton2022unsupervised}   &  0.01         &  33.27        &  27.41        &  0.02         &  20.18        &  29.84           \\
\textbf{Ours}                           &  88.93 (94\%) &  88.27 (94\%) &  86.82 (93\%) &  86.94 (93\%) &  78.20 (86\%) &  84.70 (92\%)    \\
\bottomrule
\end{tabular}
\vskip 5pt plus 1fil
  \caption{\small{
    \textbf{Main Results.} We list the segmentation results on four datasets: OpenEDS-19, OpenEDS-20, NVGaze-AR and GIW. 
    Numbers in the table represent the average Intersection over Union (IoU) for each corresponding category (outside the parentheses) and its percentage over the performance of supervised benchmark (inside the parentheses), respectively.
    The percent symbol (\%) of the IoUs are omitted.
  }}
\label{tab:main}
\end{table}

\vspace{-1cm}
\subsection{Ablation Study}
We make ablation studies mainly on two modules: 
the the prior-aware learning (PAL) strategy and the indication-refinement with SAM in EI (SAMei). 
First the ``Baseline'' setting is launched: 
training the model via outputs of the PII and initial indications from EI directly. 
Then, the PAL and SAMei are progressively added to the experiments. 
Table.~\ref{tab:ablation} exhibits the results. 
Summarily, both modules can lead to stable improvements. 
For the prior-aware learning strategy (PAL), enhancement of nearly 2\% in IoU can be observed on pupil and the iris can also acquire obvious boost. 
We attribute this primarily to its ability to leverage prior knowledge of the eye-region structure, 
thereby mitigating some of the extreme training noise in the indications. 
In the task of eye-segmentation, the SAMei improves the IoU by over 20\% in OpenEDS-19/20 and NVGaze-AR and nearly 10\% in the GIW. 
In our opinion, this huge contribution comes from the strong semantic abilities of the foundation model SAM, 
which plays an important role in distinguishing the complex and ever-changing eyelids. 
%
%
\begin{table}[]
\scriptsize
\centering
\begin{tabular}{l|ccc|ccc}
\toprule
                            & Pupil         & Iris          & Eye           & Pupil     & Iris        & Eye            \\ \hline
\makecell{\textbf{Setting}} & \multicolumn{3}{c|}{\textbf{OpenEDS-19}}      & \multicolumn{3}{c}{\textbf{OpenEDS-20}}  \\ \hline
Baseline                    &  82.40        &  77.76        &  58.77        &  85.47    &  83.93      & 65.80          \\
Baseline+PAL                &  82.64        &  79.30        &  60.77        &  87.48    &  84.89      & 63.21          \\
Baseline+SAMei              &  82.12        &  78.76        &  78.34        &  86.96    &  84.59      & 79.23          \\
Baseline+SAMei+PAL          &  83.05        &  82.62        &  80.48        &  88.32    &  87.06      & 81.43          \\ \hline
                            & \multicolumn{3}{c|}{\textbf{NVGaze-AR}}       & \multicolumn{3}{c}{\textbf{GIW}}         \\ \hline
Baseline                    &  81.67        &  84.65        &  65.16        &  85.65    &  76.98      & 74.98          \\
Baseline+PAL                &  83.98        &  86.45        &  65.77        &  88.03    &  77.93      & 75.34          \\
Baseline+SAMei              &  86.82        &  87.25        &  84.72        &  84.12    &  74.91      & 81.83          \\
Baseline+SAMei+PAL          &  88.93        &  88.27        &  86.82        &  86.94    &  78.20      & 84.70          \\
\bottomrule
\end{tabular}
\vskip 5pt plus 1fil
  \caption{\small{
    \textbf{Ablation Study.} 
    We make ablation experiments of the prior-aware learning strategy (PAL) and the indication-refinement with SAM (SAMei) in EI. 
    The IoU results are reported with all percent symbols (\%) omitted. 
  }}
\label{tab:ablation}
\end{table}

We make further discussions about the role of SAM in our work in Table.~\ref{tab:sam}. 
Under the setting of SAMei$^\dagger$, the response of SAM ($\mathcal{O}$ in Sec.~\ref{sec:ei}) is directly used for training. 
As shown in the second line of Table.~\ref{tab:sam}, a huge drop of performance can be observed. 
It proves that the original output of SAM is not reliable and our method of boundary-filtering is essential. 
Under the setting of SAMpii (the third line of Table.~\ref{tab:sam}), 
we further add SAM into PII and conduct a similar interaction and training strategy as those in the EI. 
Experiments show that the performance of the SAMpii-assisted approach is comparable with ours. 
It suggests that for pupil and iris, the image-level feature and prior knowledges are strong enough and the assistance from SAM becomes not that necessary. 
In other words, if the pretrained foundation model is not available, 
we can still obtain a segmenter with acceptable performances on pupil and iris. 
In the last line of Table.~\ref{tab:sam}, 
the roughly located starting point $p_o$ in Sec.~\ref{sec:pii} is replaced with the pupil center from the ground truth. 
It can be seen that the segmentation performances are consistent, 
which suggests that our method is not significantly influenced by the pupil location quality. 
%
%
\begin{table}[]
\centering
\begin{tabular}{l|ccc}
\toprule
                         &  Pupil        &  Iris         &  Eye     \\ \hline
Ours                     &  88.32        &  87.06        &  81.43   \\
Ours w. SAMei$^\dagger$  &  82.02        &  84.90        &  76.52   \\
Ours + SAMpii            &  89.83        &  86.45        &  81.39   \\
Ours w. pupil-gt         &  88.47        &  86.76        &  80.92   \\
\bottomrule
\end{tabular}
\vskip 5pt plus 1fil
  \caption{\small{
    \textbf{Discussions on SAM and Pupil Location.} All experiments are conducted on OpenEDS-20. 
    In ``Ours w. SAMei$^\dagger$'', the SAM outputs are directly used as training targets. 
    In ``Ours w. SAMpii'', we add SAM to PII. 
    In ``Ours w. pupil-gt'', pupil center from the ground-truth is directly used as $p_o$ during the entire training. 
  }}
\label{tab:sam}
\end{table}

\vspace{-1cm}
\subsection{Robustness over Hyper-Parameters}
As previously stated, hyper-parameters in our method are kept unchanged across all datasets in the experiments, avoiding domain-specific adjustment. 
In this section, we make a step further to validate the robustness over the key hyper-parameters. 
Experiments are conducted on OpenEDS-20. 
As illustrated in Table.~\ref{tab:robust}, we set various values for four parameters: 
the size of sliding window $k$ and $k^*$, the size $\epsilon$ of the boundary neighborhood and the threshold of the second-order derivative $th_d$. 
Segmentation performances corresponding to each setting are reported. 
To sum up, the comparison results indicate that our method is not sensitive to these hyper-parameters. 
As a matter of fact, in our design, we consistently avoid the attempts of seeking \textit{precise boundaries} in the indication process
as determining these boundaries requires accurate thresholds. 
Instead, we provide the clues in a conservative manner and leave the task of identifying precise boundary to the semantic abilities of neural network. 
We make more discussions in the Appendix. 
%
%
\vspace{-0.5cm}
\begin{table}[]
\scriptsize
\centering
\begin{tabular}{c|ccc|c|ccc}
\toprule
size of $k$ & Pupil   & Iris   & Eye   &  size of $k^*$  & Pupil  & Iris & Eye \\ \hline
$8\times{8}$    &  88.02        &  87.45        &     80.92    &  $3\times{3}$  &    88.02    &   87.01     & 81.33     \\
$9\times{9}$    &  87.95        &  87.22        &     80.97    &  $4\times{4}$  &    87.12    &   87.25     & 81.01     \\
$12\times{12}$  &  88.52        &  86.99        &     81.40    &  $7\times{7}$  &    87.55    &   87.20     & 80.98     \\ \hline
$\epsilon$ &    &    &    &  $th_d$  &   &  &  \\ \hline
25              &  87.79        &  87.25        &     81.24    &  15            &    88.10    &   86.88     & 81.21     \\
28              &  88.02        &  87.15        &     81.01    &  18            &    88.34    &   87.08     & 81.52     \\
35              &  88.23        &  86.92        &     81.35    &  25            &    88.10    &   87.38     & 81.38     \\
\bottomrule
\end{tabular}
\vskip 5pt plus 1fil
  \caption{\small{
    \textbf{Robustness over Hyper-Parameters.} We validate four main hyper-parameters: the size of $k$ and $k^*$, the size $\epsilon$ of the boundary neighborhood and the second-order derivative threshold $th_d$. All experiments are made on OpenEDS-20. 
  }}
\label{tab:robust}
\end{table}
%

%
\subsection{Visual Comparison}
\begin{figure}[t]
  \centering
  \includegraphics[width=1.\linewidth]{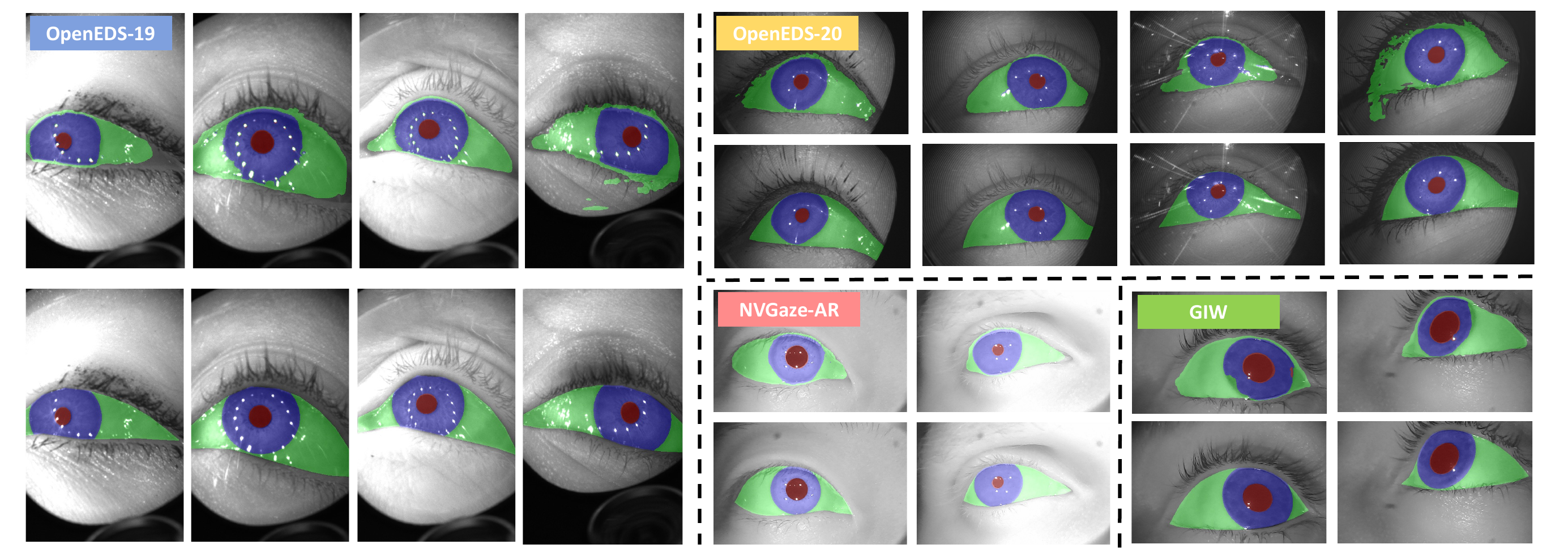}
  \vspace{-1.5\baselineskip}
  \caption{{\bf{Visual Comparisons.}}
            Visualizations are exhibited on images from the test set of OpenEDS-19, OpenEDS-20, NVGaze-AR and GIW respectively. 
            For each sample, the upper image shows the prediction while the lower image shows the ground-truth. 
            Note the eye mask is partially overlaid by the pupil/iris mask. 
          }
\label{fig:vc}
\end{figure}

We visually compare the prediction of our method with the ground-truth in Fig.~\ref{fig:vc}. 
Despite the absence of annotation, 
our approach can yield reasonable segmentation results, especially the most important category of pupil. 
The prediction errors often appear in the eye mask since distinguishing the eyelids is much more challenging. 
We provide more visual results in the Appendix. 

\section{Conclusion}
In this paper, we propose an unsupervised solution to the eye-region segmentation task for image-based eye tracking. 
First, image gradients and prior knowledges are fully used to make indications of the pupil, iris and eye. 
These clues are then leveraged to inspire a segmentation network in making dense and accruate prediction. 
Furthermore, we make use of SAM automatically to refine the indications and design the progressive learning process in an end-to-end manner. 
We think our method is a pratical solution when facing frequent updates of hardware prototype with limited labeling resources. 
%
%
%

%
%
\bibliographystyle{splncs04}
\bibliography{main}

\newpage
\setcounter{figure}{0}
\setcounter{table}{0}
\renewcommand{\thefigure}{A\arabic{figure}}  
\renewcommand{\thetable}{A\arabic{table}}  
\subsubsection{Extra Experiments of the Hyper-Parameters.}
We make extra experiments of two hyper-parameters in our method: 
the width $\tau$ of the rectangular pulse function $R_{\tau}(\cdot)$ in PII and the GSTD threshold $th_{std}$ in EI. 
As shown in Table.~\ref{tab:extra}, minor variations in these factors do not significantly affect the results. 
This observation is consistent to the declaration in the main paper that our design does not rely on accurate thesholds to indicate precise boundaries.  
Moreover, we do think there exists a reasonable range of these parameters, 
especially the size $k/k^*$ of the sliding window. 
In our opinion, this range should be highly related to the area ratio of eye over the whole image, 
which is relatively stable in near-infared eye images captured by the head-mounted devices. 
In the last two lines of Tabel.~\ref{tab:extra}, we set extreme values of $k$ and $k^*$ respectively. 
It can be observed that the performances drop significantly.
%
%
\begin{table}[]
\scriptsize
\centering
\begin{tabular}{c|ccc|c|ccc}
\toprule
$\tau$ & Pupil         & Iris          & Eye          &  $th_{std}$  & Pupil  & Iris & Eye \\ \hline
2      &  87.93        &  87.52        &     81.35    &  3           &    87.88    &   86.98     & 81.32     \\
4      &  87.89        &  88.01        &     81.17    &  7           &    88.14    &   87.01     & 80.91     \\
6      &  88.10        &  87.13        &     80.88    &  8           &    87.74    &   87.62     & 81.20     \\ \hline
size of $k$       &               &               &              &  size of $k^*$  &   &  &  \\ \hline
$3\times{3}$      &  68.22        &  62.15        &     71.01    &  $2\times{2}$   &    86.55    &   86.08     & 75.55     \\
$20\times{20}$    &  72.71        &  61.20        &     68.04    &  $15\times{15}$ &    86.32    &   85.44     & 73.97     \\
\bottomrule
\end{tabular}
\vskip 5pt plus 1fil
  \caption{\small{
    \textbf{Extra Experiments of the Hyper-Parameters.} We further study the robustness over $\tau$ in PII and $th_{std}$ in EI. 
    Then, trials of extreme values on the size of $k$ and $k^*$ are made. 
    Experiments are conducted on OpenEDS-20. 
  }}
\label{tab:extra}
\end{table}
\vspace{-1.2cm}
\subsubsection{Sparse Indications to Dense Predictions.} 
\begin{figure}[t]
  \centering
  \includegraphics[width=1.\linewidth]{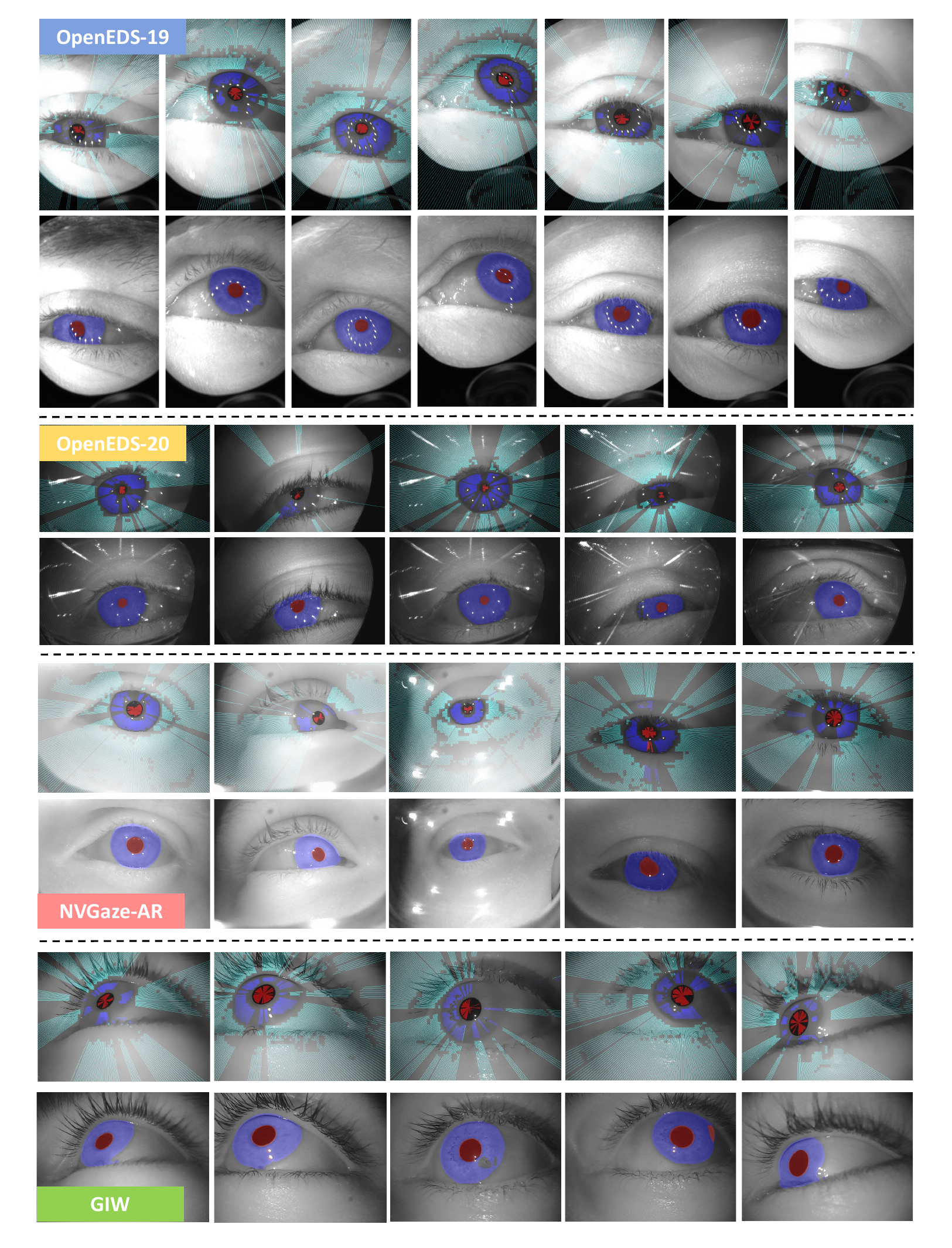}
  \vspace{-1.5\baselineskip}
  \caption{{\bf{Indications \textit{vs.} Predictions for Pupil and Iris.}}
            We visually compare the sparse indications (the upper image of each pair) and dense predictions (the lower image of each pair) for pupil (red) and iris (dark blue). 
            Pixels marked in light blue denote the background samples. 
            All indication maps come from the last update of PII outputs during training. 
            Images from the train set of OpenEDS-2019, OpenEDS-2020, NVGaze-AR and GIW are exhibited respectively. 
          }
\label{fig:pii_pred}
\end{figure}

\begin{figure}[t]
  \centering
  \includegraphics[width=1.\linewidth]{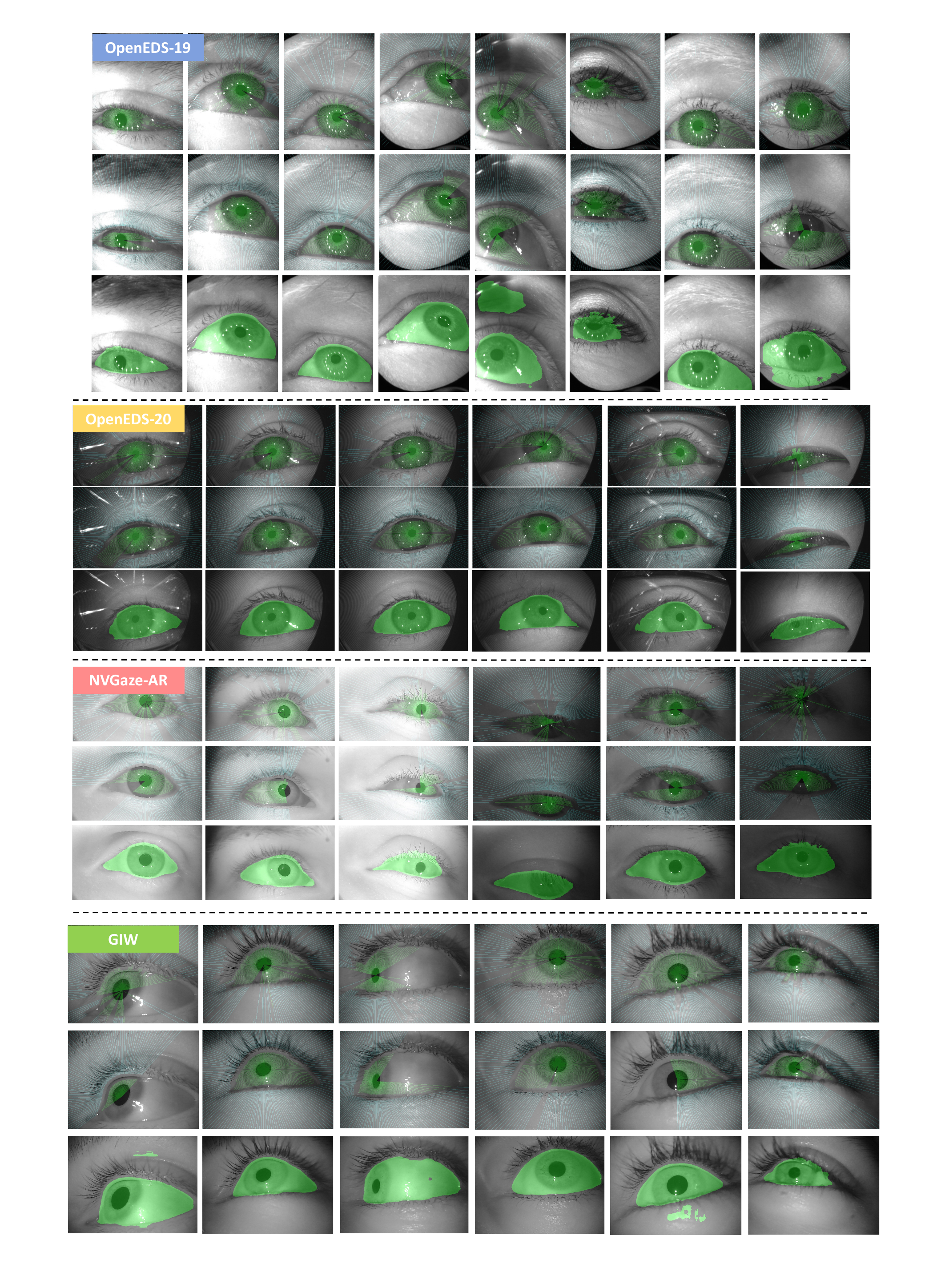}
  \vspace{-1.5\baselineskip}
  \caption{{\bf{Indications \textit{vs.} Predictions for Eye.}}
            We visually compare the sparse indications and dense predictions for eye. 
            For each example, the upper image shows the initial indications, the middle image shows the refined indications by leveraging SAM and the lower image shows the model outputs. 
            All indications come from the last update of EI outputs during training. 
            On the indication map, foreground samples are marked in green and background samples are marked in light blue. 
            Images from the train set of OpenEDS-2019, OpenEDS-2020, NVGaze-AR and GIW are exhibited respectively. 
          }
\label{fig:ei_pred}
\end{figure}

In Fig.~\ref{fig:pii_pred} and Fig.~\ref{fig:ei_pred}, 
we make visual comparisons between the indications (from the indicator PII and EI) and the predictions (from the segmentation network after training). 
In summary, it is evident that the segmentation network has the capability to draw inspiration from sparse clues to generate dense and accurate predictions, 
although these clues are limited and contain inevitable errors. 
Furthermore, our method of leveraging SAM in EI is effective, 
from which the initial eye indications can be well refined (compare the first two lines for each dataset in Fig.~\ref{fig:ei_pred}). 
However, there still exist bad cases. 
First, iris prediction is susceptible to interference from eyelashes when the iris is visually truncated by the eyelids, 
resulting in a rough and jagged mask boundary.
Then, the eye predictions are far from perfect. 
In addition to the boundaries not being smooth enough, many predictions also exhibit noticeable errors or omissions. 
We will try to alleviate these issues in our future work.

%
%
%

%
%

\end{document}